\documentclass{article}
\usepackage{iclr2027_conference,times}

\usepackage{amsmath,amsfonts,bm}

\def\eqref#1{equation~\ref{#1}}

\def\1{\bm{1}}

\DeclareMathAlphabet{\mathsfit}{\encodingdefault}{\sfdefault}{m}{sl}
\SetMathAlphabet{\mathsfit}{bold}{\encodingdefault}{\sfdefault}{bx}{n}

\usepackage{hyperref}
\usepackage{url}
\usepackage{graphicx}
\usepackage{fvextra}
\usepackage[most]{tcolorbox}
\usepackage{booktabs}
\usepackage{array}
\usepackage{threeparttable}
\usepackage{amsmath}
\usepackage{amssymb}
\usepackage{microtype}
\usepackage{xcolor}
\usepackage{colortbl}
\usepackage{multirow}
\usepackage{enumitem}
\usepackage{makecell}
\usepackage{subcaption}
\usepackage{xspace}
\usepackage{tikz}
\usetikzlibrary{arrows.meta,positioning}
\usepackage[capitalise,noabbrev]{cleveref}

\definecolor{promptaccent}{HTML}{68517E}
\newtcolorbox{promptbox}[1]{
  enhanced, breakable, colback=black!1, colframe=promptaccent!45!white,
  colbacktitle=promptaccent!10!white, coltitle=promptaccent,
  fonttitle=\small\sffamily\bfseries, title={#1},
  fontupper=\small, boxrule=0.5pt, arc=2pt,
  left=9pt, right=9pt, top=7pt, bottom=7pt,
  before skip=8pt, after skip=10pt,
  before upper={\setlength{\parindent}{0pt}\setlength{\parskip}{3pt}\raggedright}
}

\newcommand{\method}{PQ-OPD\xspace}
\newcommand{\methodfull}{Perception-Question On-Policy Distillation\xspace}

\title{Teaching What to Look For:\\
Privileged Perception Questions for Agentic\\
On-Policy Distillation in Video Captioning}

\makeatletter
\author{%
Yifan Xu$^{1,2*}$, Ruopeng Gao$^{1}$, Zhixiao Wang$^{1}$, Jiafu Tang$^{1,2*}$, Yichun Yang$^{1}$,\\
\textbf{Zihao Wang$^{2}$, Yuanxing Zhang$^{2}$, Limin Wang$^{1\dagger}$}\\[4pt]
$^{1}$Nanjing University\\
$^{2}$KlingAI Research%
\g@addto@macro\@thanks{%
  \footnotetext[1]{This work was conducted during the authors' internship at KlingAI.}%
  \footnotetext[2]{Corresponding author: Limin Wang.}%
}%
}
\makeatother

\iclrfinalcopy

\begin{document}

\maketitle

\begin{abstract}
Detailed video captioning requires comprehensive and faithful descriptions of
objects, actions, and temporal events. Fixed-frame sampling, however, can miss
brief motions, small objects, and other fine-grained evidence, motivating
captioning agents that actively seek relevant moments and inspect local regions.
Training such behavior is challenging because, unlike video question answering,
generic captioning instructions do not specify what visual evidence should be
acquired. We propose \methodfull{} (\method{}), which uses training-only
Perception Questions (PQs) to guide active video perception. PQs specify
what should be inspected without revealing the corresponding
answers, encouraging the teacher to explore the video more comprehensively
and focus on relevant visual evidence. Along trajectories generated by
the student, on-policy distillation transfers this guidance to the student's
reasoning, tool calls, and captions; no PQs are required at inference. Across
three detailed-captioning and two caption-to-QA benchmarks, \method{}
consistently improves the corresponding untrained agents across teacher--student
scales and input-frame budgets. Under our evaluation protocol, \method{}
achieves the strongest primary-metric results among the compared open-source
models on DREAM-1K, CaReBench, ShortVidBench, MotionBench, and VidCapBench-AE.
Our ablations, together with analysis of temporal coverage and answer leakage,
further indicates that the student internalizes the training-time guidance as
a more effective video exploration policy.

\end{abstract}

\section{Introduction}
\label{sec:intro}

Detailed video captioning requires comprehensive and faithful descriptions of
objects, actions, and events over time~\citep{tarsier,auroracap}. Modern
benchmarks make these requirements explicit~\citep{tarsier,carebench,auroracap,timechatcaptioner,youcook2,activitynet}:
DREAM-1K~\citep{tarsier} evaluates the precision and recall of described
events, while CaReBench~\citep{carebench} separately evaluates dynamic
actions and static objects. Recent progress has come from large-scale
captioning data and temporal alignment~\citep{tarsier2}, efficient
architectures~\citep{auroracap}, QA-based caption rewards~\citep{caprl2025,caprlpp2026}, and structured spatio-temporal supervision~\citep{percepcap2026}. These approaches, however, still perceive through
fixed sets of sampled frames. These approaches still rely on frames
selected before generation, which can miss brief motions, small objects, and precise event boundaries.

Active visual perception offers a natural alternative. Recent methods allow
video models to seek relevant temporal intervals, revisit moments, and inspect
local regions~\citep{videozoomer2026,activevideoperception2026,lenswalk2026}.
Tool-using video agents have shown the effectiveness of such interaction for
video question answering and reasoning~\citep{videoagent2024,fan2025tool,videothinker2026,thinkingvideos2026,
act2see2025,omniagentic2025,videoseeker2025,weaver2026}.
Unlike video question answering, where the question itself points to relevant
visual evidence, captioning is typically guided only by a generic instruction
such as ``describe the video in detail.'' Such an instruction specifies the
desired output but provides little guidance on what visual evidence should be
attended to. Effective visual evidence exploration is therefore crucial for
agentic captioning. This raises a natural question:
\emph{can a captioning agent learn an effective policy for deciding what visual evidence to acquire under such weakly specified instructions?}

\begin{figure}[t]
\centering
\includegraphics[width=\linewidth]{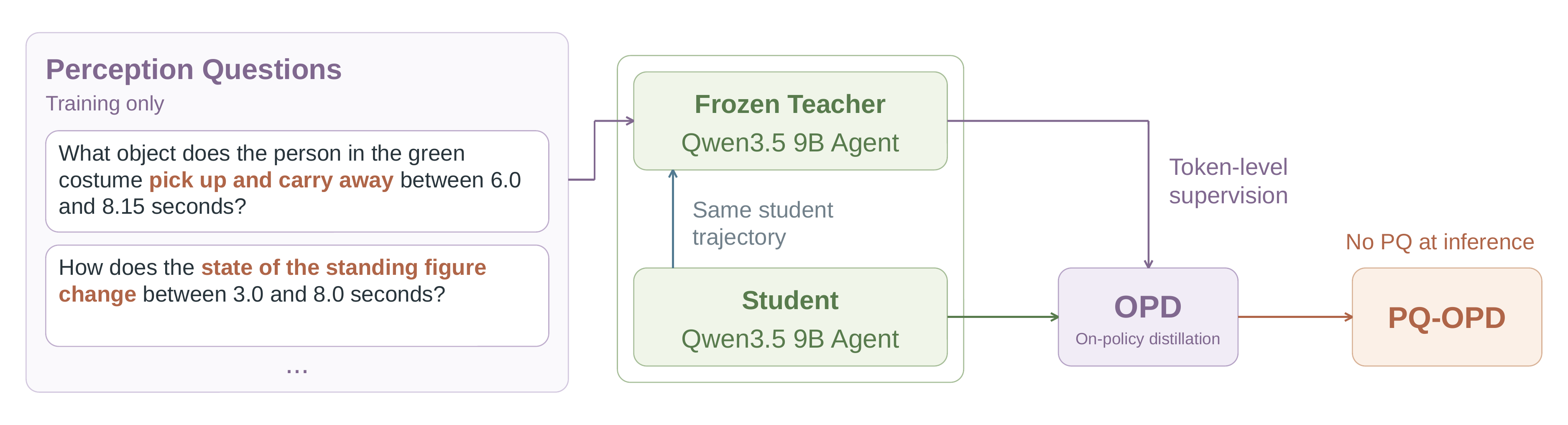}
\caption{\textbf{Training with privileged Perception Questions.}
The student generates its own tool-interaction trajectory, while a frozen
teacher receives additional Perception Questions and provides token-level
supervision. On-policy distillation transfers
this guidance to the student, which receives no PQs at inference. The
diagram illustrates the 9B teacher--student training.}
\label{fig:overview}
\end{figure}

One way to reduce this ambiguity is to make the observation targets explicit.
We find that providing \emph{Perception Questions} (PQs) encourages more
active visual evidence exploration. For example, a PQ may ask ``Which object
is the person using between 2 and 4 seconds?'' specifying what and when to
inspect without revealing the answer.
However, constructing PQs introduces an additional annotation or generation
step for each video, making them undesirable as an inference-time dependency.
We therefore aim to train the captioning agent to learn an effective visual
exploration policy from PQ-guided supervision, while requiring only the
generic captioning instruction at inference.

To this end, we propose \textbf{\methodfull{} (\method)}, illustrated in
\cref{fig:overview}, based on on-policy distillation (OPD). The student
generates its own trajectory from the generic captioning instruction, video
duration, and tool specifications, acquiring evidence through its
\texttt{seek}/\texttt{zoom} calls. A frozen teacher evaluates the same
student-generated prefixes with PQs added to its context, enabling dense
token-level supervision without a separate teacher
rollout~\citep{agarwal2024onpolicy}. A top-k approximation to Jensen–Shannon
divergence (JSD) then supervises the student’s reasoning, tool calls, and final caption. Since
observation decisions are themselves generated tokens, the same objective 
trains both evidence acquisition and caption generation. Related
self-distillation methods likewise show that students can benefit from
training-only teacher context~\citep{opsd2026,sdpo2026,viopd2024}.

We train \method{} on 5{,}979 videos with 43{,}477 PQs and conduct PQ-free evaluation on
three detailed video captioning benchmarks and two caption-to-QA benchmarks:
CaReBench, DREAM-1K, VidCapBench-AE, ShortVidBench, and MotionBench.
\method{} consistently improves over the untrained Qwen3.5-9B agent across
teacher--student scales and input-frame budgets. Under our evaluation settings,
PQ-OPD achieves the strongest results among the compared open-source models
on the primary metrics of all five benchmarks.
We further conduct ablations to analyze 
PQ-OPD. The results show that training-time guidance about what to inspect can
be learned by the captioning model and benefiting both tool-enabled agents
and fixed-frame captioners.

Our contributions are summarized as follows:
\begin{itemize}[leftmargin=*,itemsep=2pt]
\item We introduce \textbf{\methodfull{} (\method)}, an on-policy
distillation framework that uses training-only Perception Questions to
teach a video captioning agent what visual evidence to acquire through its
own tool interactions.
\item We construct 43{,}477 Perception Questions over 5{,}979 videos,
covering objects, actions, and temporal developments, providing answer-free
guidance for \method training.
\item We demonstrate consistent improvements across three detailed
captioning and two caption-to-QA benchmarks, multiple teacher--student
scales, and input-frame budgets, and show that \method benefits both caption
generation and interactive evidence acquisition.
\end{itemize}

\section{Related Work}
\label{sec}
\paragraph{Detailed video captioning.}
Recent work on detailed video captioning has improved training data, temporal
modeling, and fine-grained evaluation~\citep{tarsier,tarsier2,auroracap,carebench,meng2026videocapr1}. Beyond supervised
training, CapRL and CapRL++~\citep{caprl2025,caprlpp2026} optimize descriptions through their utility for
visual question answering, while PercepCap~\citep{percepcap2026} explicitly models spatio-temporal
perception before caption generation. These approaches operate on
visual inputs sampled before generation. \method{} instead equips the
captioner with \texttt{seek}/\texttt{zoom} tools and learns how to acquire
visual observations during generation.
\paragraph{On-policy distillation.}
On-policy distillation supervises student-generated trajectories with dense
token-level teacher distributions~\citep{agarwal2024onpolicy}. Recent work~\citep{opsd2026,sdpo2026,rethinkingopd2026,turnopd2025,exopd2026}
extends this paradigm through self-distillation, alternative objectives, and
long-horizon training strategies. In
multimodal settings, privileged teacher information has further been used to
transfer fine-grained visual perception or structured feedback~\citep{viopd2024,vicur2025,dopd2025,vaopd2025,visd2025,streamopd2025}.
Our method applies on-policy distillation to video captioning,
where a PQ-conditioned teacher provides token-level supervision along
student-controlled video interactions.
\paragraph{Video agents.}
Recent video agents~\citep{videoagent2024,fan2025tool,videozoomer2026,videothinker2026,
activevideoperception2026,lenswalk2026,thinkingvideos2026,act2see2025,
omniagentic2025,videoseeker2025,weaver2026}
actively retrieve, focus, or refine visual evidence
through iterative observation and tool use. These methods let the model
adaptively select temporal intervals, revisit relevant moments, or inspect
local regions instead of relying on a fixed visual input. Many further learn
when and what to observe from interaction trajectories, supervised
fine-tuning, or reinforcement learning. \method{} extends this direction to
open-ended video description through dense supervision from a PQ-conditioned
teacher along the student's own interaction trajectories.

\section{Method}
\label{sec:method}

\subsection{Preliminaries}
\label{sec:preliminaries}

In this section, we start with a brief review of relevant preliminaries.
Let $\mathcal{D}$ denote the input distribution, and let $p_\theta$ and
$q_\phi$ denote the student and teacher distributions, respectively. For an
input $x$ and an output sequence $y=(y_1,\ldots,y_T)$, the causal history at
token $t$ is $h_t=(x,y_{<t})$.

\paragraph{Off-Policy Distillation.}
Knowledge distillation transfers knowledge from a teacher model to a student
by training the student to reproduce the teacher's outputs~\citep{hinton2015distilling}. For language models, a common approach first
collects responses generated by the teacher and then trains the student on
these responses through supervised fine-tuning (SFT). At each position, the
student is conditioned on the teacher-generated prefix $h_t=(x,y_{<t})$ and
learns to predict the teacher's next token $y_t$. The objective is
\begin{equation}
  \mathcal{L}_{\mathrm{SFT}}(\theta)=
  -\mathbb{E}_{x\sim\mathcal{D},\,y\sim q_\phi(\cdot\mid x)}
  \left[\frac{1}{T}\sum\nolimits_{t=1}^{T}\log p_\theta(y_t\mid h_t)\right].
  \label{eq:teacher_sft}
\end{equation}
With access to the teacher's token probabilities, the student can match
the teacher's next-token distribution at each position of these fixed
responses. Writing $\mu(\cdot\mid x)$ for the fixed sequence distribution,
this token-level distillation objective is
\begin{equation}
  \mathcal{L}_{\mathrm{off}}(\theta)=
  \mathbb{E}_{x\sim\mathcal{D},\,y\sim\mu(\cdot\mid x)}
  \left[\frac{1}{T}\sum\nolimits_{t=1}^{T}
    D\!\left(q_\phi(\cdot\mid h_t),p_\theta(\cdot\mid h_t)\right)
  \right],
  \label{eq:off_policy_distillation}
\end{equation}
where $D$ measures divergence between the teacher and student next-token
distributions. Since the training sequences and their prefixes are drawn
from a fixed data source independent of the current student, the objective
is off-policy~\citep{agarwal2024onpolicy}.

\paragraph{On-Policy Distillation.}
On-policy distillation lets the student generate its own responses and uses
the teacher to provide supervision along these responses~\citep{agarwal2024onpolicy,DBLP:conf/iclr/Gu0WH24}. For each student-generated
prefix $h_t$, the teacher produces a next-token distribution, and the student
learns to match this distribution. This provides a dense training signal at
each position of the student's own sequence. The objective is
\begin{equation}
  \mathcal{L}_{\mathrm{on}}(\theta)=
  \mathbb{E}_{x\sim\mathcal{D},\,y\sim p_{\bar\theta}(\cdot\mid x)}
  \left[\frac{1}{T}\sum\nolimits_{t=1}^{T}
    D\!\left(q_\phi(\cdot\mid h_t),p_\theta(\cdot\mid h_t)\right)
  \right],
  \label{eq:on_policy_distillation}
\end{equation}
where $\bar\theta$ denotes the current student parameters used to collect
the sequences; the sampled histories are held fixed during each update.
At each training step, the current student generates a new batch of
responses for teacher supervision. During autoregressive generation, each predicted token
becomes part of the context for subsequent predictions. Training on
student-generated responses therefore allows the teacher to guide the
continuation of sequences that the student itself produces~\citep{agarwal2024onpolicy}. We use symmetric JSD as the divergence in our
method, as detailed below.

\subsection{Perception-Question On-Policy Distillation}
\label{sec:pq_opd}
\label{sec:pq_privilege}
\label{sec:agentic_opd}

\paragraph{Agentic video captioning.}

Let $V$ be a video and $x$ the input prompt including task instructions and tool
specifications; the task is to produce a detailed caption of $V$, which the
model emits as $y$. At interaction round $r$ it may issue tool calls:
\texttt{seek}$(t_{\rm start},t_{\rm end})$ returns up to eight frames sampled
uniformly from a time interval, and \texttt{zoom}$(t,\mathbf{b})$ returns a
single spatial crop at time $t$, with $\mathbf{b}$ a normalized xyxy box in
$[0,1000]$. Inspection ends when the model emits $y$ or when it reaches the
round limit, and the returned images come from an environment $\mathcal{E}$,
so we write a trajectory as
\begin{equation}
  \tau=(x,\,z_1,a_1,o_1,\ldots,z_R,a_R,o_R,\,z_f,y),
\end{equation}
where $R$ is the number of tool-interaction rounds, $z_r$ is the reasoning
at round $r$, $a_r$ denotes tool calls issued at round $r$, and
$o_r=\mathcal{E}(V,a_r)$ the returned observations. The final reasoning
$z_f$ precedes the caption $y$.

A single autoregressive model generates $z_r$, $a_r$, and $y$ from one
distribution. We call the part of its behavior that decides what to inspect
next and when to stop the \emph{exploration policy}.

\paragraph{Perception-question guidance.}
For each training video $V$, we construct a set of Perception Questions
$\mathcal{Q}(V)=\{q_1,\ldots,q_K\}$ from its visual content. These
questions direct attention to visual aspects relevant to detailed captioning,
including main objects, actions and interactions, and temporal developments
such as action sequences, state changes, and repetition. Each question
specifies what should be inspected and the temporal interval in which it can
be resolved, while leaving the corresponding visual evidence unspecified, as
illustrated in \cref{fig:overview}. For
example, a question may ask ``Which object is the person using between 2 and
4 seconds?'' without revealing the identity of the object. Taken together,
the questions guide the model toward relevant
visual evidence without specifying what content the video contains.

Notably, we do not provide PQ answers during training, as our purpose
is to improve visual exploration rather than supply answers through text. We expose
only what to inspect and when, while withholding the corresponding answers. Revealing the answers can create
a textual shortcut, leading to less active visual exploration and fewer tool
calls. This design is supported by prior work~\citep{vicur2025,rge} and our
answer-leakage study in \cref{sec:analysis}, where providing PQ answers reduces
the student's final caption quality.

The teacher and student are initialized from their respective pretrained
Qwen3.5 checkpoints, and the teacher remains frozen throughout training.
In the same-scale 9B $\rightarrow$ 9B setting, they share the same pretrained
initialization; in cross-scale settings, they use checkpoints of different
model sizes.
For each student-generated trajectory $\tau$, the student
is conditioned on its causal history $h_t$, while the teacher receives the
same history augmented with $\mathcal{Q}(V)$
(\cref{app:teacher_prompt} gives the exact template). At each assistant token
$t$, they produce the next-token distributions
$p_\theta(\cdot\mid h_t)$ and
$q_{\phi_0}(\cdot\mid h_t,\mathcal{Q}(V))$, respectively. The teacher thus
provides token-level supervision along the trajectory generated by the
student, as illustrated in \cref{fig:overview}.

\paragraph{On-policy training objective.}
At each training step, the student generates a batch of trajectories
$\mathcal{B}=\{(V_i,x_i,\tau_i)\}_{i=1}^{B}$, with
$\tau_i\sim P_{\bar\theta}(\cdot\mid V_i,x_i)$. Here $P_{\bar\theta}$ is
the trajectory distribution induced by the student with sampling parameters
$\bar\theta$ interacting with the video environment. The teacher provides
token-level supervision along these trajectories. We minimize the
token-averaged batch loss
\begin{equation}
  \widehat{\mathcal{L}}(\theta;\mathcal{B})=
  \frac{\sum\nolimits_{i=1}^{B}
    \sum\nolimits_{t\in\mathcal{A}(\tau_i)}
    D_{\mathrm{JS}}^{(k)}\!\left(
      q_{\phi_0}(\cdot\mid h_{i,t},\mathcal{Q}(V_i)),
      p_\theta(\cdot\mid h_{i,t})\right)}
  {\sum\nolimits_{i=1}^{B}|\mathcal{A}(\tau_i)|},
  \label{eq:opd}
\end{equation}
where $h_{i,t}$ is the causal history before assistant token $t$ in
trajectory $\tau_i$, including the prompt, preceding assistant tokens, and
tool observations returned so far. The set $\mathcal{A}(\tau_i)$ contains
all assistant-token positions across reasoning, tool calls, and the final
caption. $D_{\mathrm{JS}}^{(k)}$ denotes symmetric JSD computed on the
teacher's top-$k$ tokens and a tail bucket containing the remaining
probability mass, with $k=100$. Gradients are computed only through the
student distribution, with sampled trajectories, observations, and teacher
distributions held fixed during each update.

\section{Experiments}
\label{sec:experiments}

\subsection{Setup}
\label{sec:setup}

\paragraph{Training data.}
We construct 43{,}477 Perception Questions for 5{,}979 training videos from
Tarsier2-Recap-585K~\citep{tarsier2} using Gemini 3.1 Pro~\citep{geminiteam2026gemini31problog}, averaging
7.27 questions per video. The prompt template is provided
in \cref{app:pq_generation_prompt}. Generated answers and evidence descriptions are retained only for
quality control and analysis and are excluded from the training. All 43{,}477
PQs pass an automatic validator covering 97.6\% of each video's duration on
average, and expert verification on 50 sampled videos (344 PQs) confirms that
99.7\% target visually present content without hallucinated premises and 99.4\%
have intervals covering the target evidence (\cref{app:pq_quality}).

\paragraph{Model training.}
Our primary setting uses Qwen3.5 9B for both student and teacher, with
the teacher frozen at initialization. We train on all 5{,}979 videos and
43{,}477 PQs for 372 optimizer steps with a peak learning rate of
$2\times10^{-6}$, while freezing the vision encoder and aligner.
We additionally train models under 27B $\rightarrow$ 9B and
9B $\rightarrow$ 4B teacher--student settings, initializing each model
from the Qwen3.5 checkpoint of its corresponding size and keeping the
teacher frozen. We approximate
full-vocabulary JSD using the teacher's top-$k$ support with $k=100$ and
aggregate the remaining probability mass into a single tail bucket.
Full training hyperparameters are provided in
\cref{app:training_details}.

\paragraph{Benchmarks and inference setting.}
We evaluate caption quality on CaReBench~\citep{carebench}, DREAM-1K~\citep{tarsier}, and VidCapBench-AE~\citep{vidcapbench}. CaReBench measures
Object and Action content, DREAM-1K evaluates event coverage with AutoDQ, and
VidCapBench-AE assesses detailed caption quality across multiple dimensions.
As the original judges used by CaReBench and DREAM-1K are no longer available
through public API services, we use GPT-4.1~\citep{openai2025gpt41} as the judge for both benchmarks;
VidCapBench-AE uses GPT-4o~\citep{openai2024gpt4o}. We further evaluate caption-to-QA performance on MotionBench and
ShortVidBench by replacing the video with the generated caption.
Gemini 3.1 Pro~\citep{geminiteam2026gemini31problog} first answers each
question from the caption, and a separate independent call evaluates the
accuracy of the resulting answer.
At inference time, the agent receives no PQs and may
use up to 16 rounds of \texttt{seek}/\texttt{zoom} calls within a
128K context and a 128 total input frames limit.
\texttt{seek} returns up to eight frames
and \texttt{zoom} returns one crop, with each image limited to 460{,}800
pixels.

\subsection{Main Results}
\label{sec:main_results}

\Cref{tab:main_results} and
\cref{tab:vidcapbench_ae_results} compare \method{} with direct video models
and the untrained Qwen3.5 9B Agent across fine-grained captioning and
caption-to-QA benchmarks. Qwen3.5 9B uses 32-frame direct video input, while
the agent variants acquire visual observations through tools.
VideoCap-R1~\citep{meng2026videocapr1} has no publicly released weights, so we reproduced it following
the training recipe in its paper and evaluated the reproduced model under the same protocol.

\begin{table}[t]
\centering
\caption{\textbf{Evaluation results on DREAM-1K~\citep{tarsier}, CaReBench~\citep{carebench}, ShortVidBench~\citep{shortvidbench}, and MotionBench~\citep{motionbench}.} For \method{}, the model size denotes the student; the teacher size is given in parentheses. Scores are reported as percentages. We highlight the \textbf{best} and \underline{second-best} available open-source results.}
\label{tab:main_results}
\fontsize{7.5}{9}\selectfont
\setlength{\tabcolsep}{3.5pt}
\renewcommand{\arraystretch}{1.12}
\resizebox{\linewidth}{!}{%
\begin{tabular}{@{}>{\raggedright\arraybackslash}m{0.275\linewidth}c|cc|c|c@{}}
\toprule
\multicolumn{1}{c}{\multirow{2}{*}{Model}} & \multicolumn{1}{c|}{DREAM-1K} & \multicolumn{2}{c|}{CaReBench} & \multicolumn{1}{c|}{ShortVidBench} & MotionBench \\
\cmidrule(l){2-6}
& \multicolumn{1}{c|}{Event F1/R/P} & Action F1/R/P & \multicolumn{1}{c|}{Object F1/R/P} & \multicolumn{1}{c|}{Acc.} & Acc. \\
\midrule
\multicolumn{6}{@{}l@{}}{\textit{Proprietary models (fixed 32 frames)}} \\
{\color[HTML]{9B9B9B} Gemini 3 Flash~\citep{gemini3flash}} & \multicolumn{1}{c|}{{\color[HTML]{9B9B9B} 37.6/48.5/30.7}} & {\color[HTML]{9B9B9B} 42.4/39.6/45.7} & \multicolumn{1}{c|}{{\color[HTML]{9B9B9B} 40.3/37.8/43.2}} & \multicolumn{1}{c|}{{\color[HTML]{9B9B9B} 78.6}} & {\color[HTML]{9B9B9B} 57.6} \\
{\color[HTML]{9B9B9B} Gemini 3.1 Pro~\citep{geminiteam2026gemini31problog}} & \multicolumn{1}{c|}{{\color[HTML]{9B9B9B} 38.2/45.2/33.1}} & {\color[HTML]{9B9B9B} 42.0/37.4/47.7} & \multicolumn{1}{c|}{{\color[HTML]{9B9B9B} 39.3/32.6/49.6}} & \multicolumn{1}{c|}{{\color[HTML]{9B9B9B} 78.4}} & {\color[HTML]{9B9B9B} 55.9} \\
\midrule
\multicolumn{6}{@{}l@{}}{\textit{Open-source models (fixed 32 frames)}} \\
Qwen2.5-VL 7B~\citep{Qwen2.5-VL} & \multicolumn{1}{c|}{27.2/27.9/26.6} & 30.9/23.6/45.0 & \multicolumn{1}{c|}{34.1/27.6/44.6} & \multicolumn{1}{c|}{70.0} & 50.7 \\
Kimi-VL A3B~\citep{kimivl} & \multicolumn{1}{c|}{27.5/32.6/23.8} & 35.1/29.0/44.4 & \multicolumn{1}{c|}{37.7/32.0/45.8} & \multicolumn{1}{c|}{74.2} & 52.1 \\
Qwen3-VL 8B~\citep{qwen3vl} & \multicolumn{1}{c|}{29.1/36.3/24.3} & 35.9/31.2/42.2 & \multicolumn{1}{c|}{38.6/35.4/42.5} & \multicolumn{1}{c|}{72.8} & 53.0 \\
Qwen3.5 9B~\citep{qwen3.5} & \multicolumn{1}{c|}{30.3/40.5/24.1} & 36.9/34.8/39.3 & \multicolumn{1}{c|}{38.3/42.1/35.1} & \multicolumn{1}{c|}{74.8} & 56.6 \\
VideoCap-R1~\citep{meng2026videocapr1} & \multicolumn{1}{c|}{28.8/32.0/26.1} & 33.5/26.6/\underline{45.2} & \multicolumn{1}{c|}{33.8/26.1/\textbf{48.0}} & \multicolumn{1}{c|}{71.1} & 53.0 \\
OwlCap-7B~\citep{DBLP:conf/aaai/ZhongHZZHLTB26} & \multicolumn{1}{c|}{27.2/28.2/26.2} & 33.7/27.3/44.0 & \multicolumn{1}{c|}{34.5/26.9/\underline{47.9}} & \multicolumn{1}{c|}{70.3} & 53.1 \\
\midrule
\multicolumn{6}{@{}l@{}}{\textit{Agent models (with seek/zoom tools)}} \\
Qwen3.5 9B Agent & \multicolumn{1}{c|}{31.3/38.9/26.1} & 40.0/\textbf{37.4}/43.0 & \multicolumn{1}{c|}{39.6/\textbf{47.1}/34.1} & \multicolumn{1}{c|}{75.0} & 57.0 \\
\textbf{\method{} 4B (9B Teacher)} & \multicolumn{1}{c|}{32.6/39.4/\textbf{27.8}} & 38.9/34.8/44.3 & \multicolumn{1}{c|}{\underline{41.4}/41.5/41.4} & \multicolumn{1}{c|}{\underline{77.0}} & 57.9 \\
\textbf{\method{} 9B (9B Teacher)} & \multicolumn{1}{c|}{\underline{33.3}/\underline{41.5}/\textbf{27.8}} & \underline{40.1}/\underline{36.6}/44.3 & \multicolumn{1}{c|}{41.0/41.6/40.5} & \multicolumn{1}{c|}{76.1} & \underline{59.0} \\
\textbf{\method{} 9B (27B Teacher)} & \multicolumn{1}{c|}{\textbf{33.4}/\textbf{42.3}/\underline{27.6}} & \textbf{40.6}/36.3/\textbf{46.0} & \multicolumn{1}{c|}{\textbf{41.7}/\underline{44.4}/39.3} & \multicolumn{1}{c|}{\textbf{78.5}} & \textbf{59.6} \\
\bottomrule
\end{tabular}%
}
\end{table}

\begin{table}[t]
\centering
\caption{\textbf{Evaluation results on VidCapBench-AE~\citep{vidcapbench}.}
Each group reports Accuracy/Precision/Coverage/Conciseness, as percentages.
For \method{}, the model size denotes the student and parentheses
specify the teacher. The \textbf{best} and \underline{second-best}
available open-source results are highlighted for each metric.}
\label{tab:vidcapbench_ae_results}
\scriptsize
\setlength{\tabcolsep}{3pt}
\renewcommand{\arraystretch}{1.12}
\resizebox{\linewidth}{!}{%
\begin{tabular}{@{}>{\raggedright\arraybackslash}m{0.25\linewidth}ccccc@{}}
\toprule
\multicolumn{1}{c}{\multirow{3}{*}{Model}} & \multicolumn{5}{c}{VidCapBench-AE} \\
\cmidrule(l){2-6}
& \multicolumn{1}{c|}{Overall} & \multicolumn{1}{c|}{Video Aesthetics} & \multicolumn{1}{c|}{Video Content} & \multicolumn{1}{c|}{Video Motion} & Physical Laws \\
\cmidrule(l){2-6}
& \multicolumn{1}{c|}{Acc./Pre./Cov./Con.} & \multicolumn{1}{c|}{Acc./Pre./Cov./Con.} & \multicolumn{1}{c|}{Acc./Pre./Cov./Con.} & \multicolumn{1}{c|}{Acc./Pre./Cov./Con.} & Acc./Pre./Cov./Con. \\
\midrule
\multicolumn{6}{@{}l@{}}{\textit{Proprietary models (fixed 32 frames)}} \\
{\color[HTML]{9B9B9B} GPT-4o~\citep{openai2024gpt4o}} & \multicolumn{1}{c|}{{\color[HTML]{9B9B9B} 16.8/57.4/86.0/5.9}} & \multicolumn{1}{c|}{{\color[HTML]{9B9B9B} 14.1/47.6/83.4/4.9}} & \multicolumn{1}{c|}{{\color[HTML]{9B9B9B} 17.5/61.7/87.2/6.1}} & \multicolumn{1}{c|}{{\color[HTML]{9B9B9B} 10.2/41.3/84.0/3.6}} & {\color[HTML]{9B9B9B} 27.9/62.1/85.4/9.7} \\
{\color[HTML]{9B9B9B} Gemini 3 Flash~\citep{gemini3flash}} & \multicolumn{1}{c|}{{\color[HTML]{9B9B9B} 19.3/58.1/90.5/7.3}} & \multicolumn{1}{c|}{{\color[HTML]{9B9B9B} 17.1/47.3/87.5/6.4}} & \multicolumn{1}{c|}{{\color[HTML]{9B9B9B} 19.6/62.8/91.7/7.4}} & \multicolumn{1}{c|}{{\color[HTML]{9B9B9B} 12.9/49.5/89.8/4.9}} & {\color[HTML]{9B9B9B} 33.9/58.4/91.7/12.8} \\
\midrule
\multicolumn{6}{@{}l@{}}{\textit{Open-source models (fixed 32 frames)}} \\
Qwen2-VL 7B~\citep{wang2024qwen2vl} & \multicolumn{1}{c|}{11.1/47.1/77.0/6.4} & \multicolumn{1}{c|}{12.4/44.3/78.7/\underline{7.2}} & \multicolumn{1}{c|}{9.9/48.3/75.9/5.7} & \multicolumn{1}{c|}{4.0/22.7/78.2/2.3} & 26.1/\textbf{59.4}/81.2/\underline{15.1} \\
Kimi-VL A3B~\citep{kimivl} & \multicolumn{1}{c|}{15.4/54.5/85.9/2.8} & \multicolumn{1}{c|}{13.5/46.9/81.1/2.4} & \multicolumn{1}{c|}{15.7/58.3/87.7/2.8} & \multicolumn{1}{c|}{6.2/36.5/85.3/1.1} & 28.1/52.0/89.4/5.1 \\
MiMo-VL~\citep{mimovl} & \multicolumn{1}{c|}{15.8/52.3/86.8/\underline{7.5}} & \multicolumn{1}{c|}{14.5/44.9/85.0/6.9} & \multicolumn{1}{c|}{15.8/55.8/87.8/\underline{7.5}} & \multicolumn{1}{c|}{8.4/36.1/84.9/\underline{4.0}} & 28.4/56.8/85.4/13.6 \\
Intern2VL 8B~\citep{intern2vl} & \multicolumn{1}{c|}{10.2/43.0/84.9/2.5} & \multicolumn{1}{c|}{9.1/36.3/84.4/2.2} & \multicolumn{1}{c|}{10.0/46.1/85.2/2.4} & \multicolumn{1}{c|}{4.4/18.0/81.3/1.1} & 23.6/52.8/85.7/5.8 \\
Qwen3-VL 8B~\citep{qwen3vl} & \multicolumn{1}{c|}{17.8/56.8/88.5/4.4} & \multicolumn{1}{c|}{14.5/47.3/84.8/3.6} & \multicolumn{1}{c|}{18.6/61.1/90.3/4.6} & \multicolumn{1}{c|}{9.3/39.2/84.0/2.3} & 32.7/57.3/87.7/8.1 \\
Qwen3.5 9B~\citep{qwen3.5} & \multicolumn{1}{c|}{17.2/52.7/88.5/5.8} & \multicolumn{1}{c|}{15.3/44.5/85.5/5.2} & \multicolumn{1}{c|}{17.4/56.7/89.6/5.9} & \multicolumn{1}{c|}{9.3/37.6/91.1/3.2} & 31.2/52.2/89.4/10.5 \\
VideoCap-R1~\citep{meng2026videocapr1} & \multicolumn{1}{c|}{14.2/50.1/84.5/\textbf{8.5}} & \multicolumn{1}{c|}{12.3/42.9/82.7/\textbf{7.4}} & \multicolumn{1}{c|}{14.1/53.3/85.0/\textbf{8.5}} & \multicolumn{1}{c|}{7.6/35.9/85.3/\textbf{4.6}} & 31.9/55.0/87.7/\textbf{19.3} \\
OwlCap-7B~\citep{DBLP:conf/aaai/ZhongHZZHLTB26} & \multicolumn{1}{c|}{15.9/52.8/84.9/6.1} & \multicolumn{1}{c|}{14.0/44.1/83.2/5.4} & \multicolumn{1}{c|}{15.9/56.8/85.5/6.1} & \multicolumn{1}{c|}{9.8/36.6/84.9/3.8} & 31.2/54.7/88.2/12.0 \\
\midrule
\multicolumn{6}{@{}l@{}}{\textit{Agent models (with seek/zoom tools)}} \\
Qwen3.5 9B Agent & \multicolumn{1}{c|}{19.9/58.2/\underline{91.4}/4.4} & \multicolumn{1}{c|}{15.3/46.7/86.1/3.4} & \multicolumn{1}{c|}{\underline{21.3}/\underline{63.5}/\underline{93.7}/4.7} & \multicolumn{1}{c|}{8.0/36.8/89.3/1.8} & 33.9/57.5/91.0/7.5 \\
\textbf{\method{} 4B (9B Teacher)} & \multicolumn{1}{c|}{19.6/57.8/90.7/4.2} & \multicolumn{1}{c|}{\underline{16.7}/47.9/85.5/3.6} & \multicolumn{1}{c|}{20.1/62.4/92.8/4.3} & \multicolumn{1}{c|}{11.1/40.0/\underline{93.3}/2.4} & \underline{36.7}/58.1/\textbf{93.5}/7.9 \\
\textbf{\method{} 9B (9B Teacher)} & \multicolumn{1}{c|}{\underline{20.6}/\underline{58.7}/91.2/4.6} & \multicolumn{1}{c|}{16.6/\underline{48.5}/\textbf{87.0}/3.7} & \multicolumn{1}{c|}{\textbf{21.5}/\underline{63.5}/92.9/4.8} & \multicolumn{1}{c|}{\textbf{13.8}/\underline{40.4}/\textbf{94.7}/3.1} & \textbf{37.2}/\underline{59.3}/92.0/8.3 \\
\textbf{\method{} 9B (27B Teacher)} & \multicolumn{1}{c|}{\textbf{20.8}/\textbf{59.6}/\textbf{91.8}/4.8} & \multicolumn{1}{c|}{\textbf{18.0}/\textbf{49.6}/\underline{86.6}/4.2} & \multicolumn{1}{c|}{\textbf{21.5}/\textbf{64.2}/\textbf{94.0}/5.0} & \multicolumn{1}{c|}{\underline{12.9}/\textbf{48.5}/89.8/3.0} & 33.4/57.2/\underline{92.2}/7.8 \\
\bottomrule
\end{tabular}%
}
\end{table}

\textbf{Fine-grained caption quality.}
Across DREAM-1K, CaReBench, and VidCapBench-AE, \method{} consistently
improves caption quality. The 27B $\rightarrow$ 9B setting
achieves the best open-source Event F1 on DREAM-1K (33.4) and the best
Action and Object F1 on CaReBench (40.6 and 41.7). On VidCapBench-AE, it
also obtains the highest open-source overall accuracy (20.8) and precision
(59.6), the highest video-aesthetics accuracy (18.0), and tied-highest
video-content accuracy (21.5) among the compared open-source models. 
The 9B $\rightarrow$ 9B setting performs particularly well
on dynamic content, reaching the best open-source motion accuracy (13.8) and
physical-law accuracy (37.2).

\textbf{Caption-to-QA evaluation.}
We further evaluate the generated captions by using them in place of the
original video input on ShortVidBench~\citep{shortvidbench} and MotionBench~\citep{motionbench}. The 27B
$\rightarrow$ 9B model achieves 78.5 accuracy on ShortVidBench and 59.6 on
MotionBench, obtaining the best open-source results on both benchmarks. It
also surpasses the listed proprietary models on MotionBench and performs
comparably on ShortVidBench.

\textbf{Distillation across model scales.}
The improvements are consistent across teacher--student scales. With a 9B
teacher and student, \method{} improves DREAM-1K Event F1 from 31.3 to 33.3,
CaReBench Object F1 from 39.6 to 41.0, ShortVidBench accuracy from 75.0 to
76.1, and MotionBench accuracy from 57.0 to 59.0. The 4B student also outperforms
the untrained 9B agent on DREAM-1K, ShortVidBench, and
MotionBench, showing that the behavior can be transferred to a
smaller student.

\subsection{Ablation Study}
\label{sec:ablation}

\begin{table}[t]
\centering
\caption{
\textbf{Ablation study of PQ guidance and training strategies on the 9B backbone.}
CaReBench F1 is the arithmetic mean of Action F1 and Object F1;
DREAM-1K F1 is event F1.
``Train PQ'' indicates that PQs are used during training, while
``Infer. PQ'' indicates that video-specific PQs are provided at inference.
Rows \#5--\#11 receive no PQs at inference; \#9 uses video-agnostic generic
questions (\cref{app:leakage_and_checklist}).
}
\label{tab:ablation}

\small
\setlength{\tabcolsep}{4pt}
\renewcommand{\arraystretch}{1.12}

\resizebox{\linewidth}{!}{%
\begin{tabular}{
@{}
>{\centering\arraybackslash}p{0.045\linewidth}
>{\raggedright\arraybackslash}p{0.455\linewidth}
>{\centering\arraybackslash}p{0.095\linewidth}
>{\centering\arraybackslash}p{0.095\linewidth}
>{\centering\arraybackslash}p{0.155\linewidth}
>{\centering\arraybackslash}p{0.155\linewidth}
@{}
}
\toprule
\# & Model Variant
& Train PQ
& Infer. PQ
& CaReBench F1
& DREAM-1K F1 \\
\midrule

\multicolumn{6}{@{}l}{
\textit{Fixed observations (no seek/zoom tools)}
} \\
\#1 & Baseline (Qwen3.5 9B)
& -- & -- & 37.6 & 30.3 \\

\#2 & \quad + PQ at inference
& -- & \checkmark & 37.4 & 33.4 \\

\#5 & \quad + PQ-OPD
& \checkmark & -- & 38.0 & 31.9 \\

\midrule

\multicolumn{6}{@{}l}{
\textit{Interactive observations (seek/zoom tools)}
} \\
\#3 & Base agent (Qwen3.5 9B)
& -- & -- & 39.8 & 31.3 \\

\#4 & \quad + PQ at inference
& -- & \checkmark & 40.6 & 33.4 \\

\#6 & \quad + PQ trace off-policy distillation (NTP Loss)
& \checkmark & -- & 39.7 & 32.6 \\

\#7 & \quad + PQ trace off-policy distillation (JSD Loss)
& \checkmark & -- & 40.2 & 32.9 \\

\#8 & \quad + Shuffled-PQ-OPD
& \checkmark & -- & 40.1 & 31.5 \\

\#9 & \quad + Generic Question OPD
& -- & -- & 40.2 & 32.8 \\

\#10 & \quad + PQs w/o timestamps OPD
& \checkmark & -- & 40.6 & 31.9 \\

\rowcolor{gray!10}
\#11 & \quad + PQ-OPD (Ours)
& \checkmark & -- & \textbf{40.6} & \textbf{33.3} \\

\bottomrule
\end{tabular}%
}
\end{table}

\textbf{Effect of PQ guidance.}
For the inference-only variants (\#2 and \#4), PQs are generated from each
CaReBench or DREAM-1K video using the same construction procedure as the
training PQs. Providing these PQs at inference substantially improves
DREAM-1K in both direct captioning and agentic settings. From \#1 to \#2,
DREAM-1K F1 increases from 30.3 to 33.4, while CaReBench remains comparable.
With \texttt{seek}/\texttt{zoom} tools, \#3 to \#4 improves DREAM-1K F1 from
31.3 to 33.4 and CaReBench F1 from 39.8 to 40.6. These results show that PQs
provide useful video-specific guidance, particularly for event-level
description.

\textbf{Effect of tool-aware on-policy training.}
PQ-OPD benefits captioning even when visual observations are fixed. Without
tools, training improves CaReBench F1 from 37.6 to 38.0 and DREAM-1K F1 from
30.3 to 31.9 (\#1 $\rightarrow$ \#5). In this setting, PQ-conditioned
supervision can improve how the student interprets and describes the given
frames, but cannot alter which visual evidence is observed.

With \texttt{seek}/\texttt{zoom} tools, the same supervision also covers the
student's observation decisions along its interaction trajectory. PQ-OPD
improves the untrained agent from 39.8 to 40.6 on CaReBench and from 31.3 to
33.3 on DREAM-1K (\#3 $\rightarrow$ \#11). This larger improvement is
consistent with PQ guidance being particularly useful in the agentic setting,
where training can shape both evidence acquisition and caption generation.
Randomly redistributing individual PQs across training videos (Shuffled-PQ,
\#8) slightly improves the untrained agent, reaching 40.1 CaReBench F1 and
31.5 DREAM-1K F1. We hypothesize that PQs shift the teacher's output
distribution to favor more exploration, consistent with the
context-induced teacher behavior observed in OP$^2$SD~\citep{ichihara2026op2sd}.
Replacing PQs with eight neutral quarter-interval questions matched in count
and type distribution (Generic Question OPD, \#9;
\cref{app:leakage_and_checklist}) reaches 40.2 CaReBench F1 and 32.8 DREAM-1K
F1, while removing timestamps from our PQs (\#10) leaves CaReBench at 40.6 but
lowers DREAM-1K to 31.9, indicating that video-specific wording guides
\emph{what} to inspect and timestamps guide \emph{when}.

We further compare on-policy distillation with off-policy distillation on PQ-conditioned teacher
traces. For PQ trace off-policy distillation (NTP Loss, \#6), we sample four teacher rollouts per training
video and fine-tune the student on the resulting interaction trajectories without PQ input, and we also evaluate matching soft top-$k$ teacher distributions on the same fixed trajectories (JSD Loss, \#7), which reaches 40.2 CaReBench F1 and 32.9 DREAM-1K F1.
Under the same tool interface, PQ-OPD (\#11) improves CaReBench F1 from 39.7
to 40.6 and DREAM-1K F1 from 32.6 to 33.3. Unlike off-policy distillation on fixed teacher
trajectories, on-policy distillation provides supervision along the
observation histories generated by the current student, making it better
suited to learning interactive perception policies.

\begin{table}[t]
\centering
\caption{PQ guidance across teacher--student configurations. Both methods
use seek/zoom tools; only PQ-OPD conditions the teacher on PQs, and all
students are evaluated without PQs. Arrows denote teacher $\rightarrow$
student sizes. Scores are percentages; CaReBench F1 averages Action and
Object F1, and DREAM-1K F1 is event F1. Bold denotes the better score
within each teacher--student pair.}
\label{tab:cross_scale}
\small
\setlength{\tabcolsep}{3pt}
\renewcommand{\arraystretch}{1.12}
\resizebox{\linewidth}{!}{%
\begin{tabular}{@{}>{\centering\arraybackslash}p{0.20\linewidth}>{\raggedright\arraybackslash}p{0.38\linewidth}>{\centering\arraybackslash}p{0.17\linewidth}>{\centering\arraybackslash}p{0.17\linewidth}@{}}
\toprule
Teacher $\rightarrow$ Student & Training & CaReBench F1 & DREAM-1K F1 \\
\midrule
27B $\rightarrow$ 9B & Vanilla OPD & 40.1 & 31.2 \\
\rowcolor{gray!10}
27B $\rightarrow$ 9B & PQ-OPD & \textbf{41.2} & \textbf{33.4} \\
\midrule
9B $\rightarrow$ 4B & Vanilla OPD & 39.4 & 30.8 \\
\rowcolor{gray!10}
9B $\rightarrow$ 4B & PQ-OPD & \textbf{40.2} & \textbf{32.6} \\
\bottomrule
\end{tabular}%
}
\end{table}

\textbf{PQ guidance across model scales.}
\Cref{tab:cross_scale} shows consistent gains from PQ-conditioned OPD across
teacher--student scales. With a 27B teacher and 9B student, PQ-OPD improves
CaReBench F1 from 40.1 to 41.2 and DREAM-1K F1 from 31.2 to 33.4. With a 9B
teacher and 4B student, the corresponding scores increase from 39.4 to 40.2
and from 30.8 to 32.6. The gains at both scales show that PQ-conditioned
supervision is not specific to a single teacher--student setting.

\subsection{Analysis}
\label{sec:analysis}

\textbf{Performance across input-frame budgets.}
A captioning agent must decide both where to inspect the video and how to use
a limited visual input. We study this behavior by varying the input-frame
budget from 2 to 64 frames for the 9B student trained with a 9B teacher. The
budget specifies the maximum number of frames that may be returned across all
tool calls. We compare uniformly sampled direct video input, the untrained
Qwen3.5~9B Agent, and \method{}, with the two agents sharing the same tools,
prompts, and frame limits.

\begin{figure}[t]
  \centering
  \includegraphics[width=\linewidth]{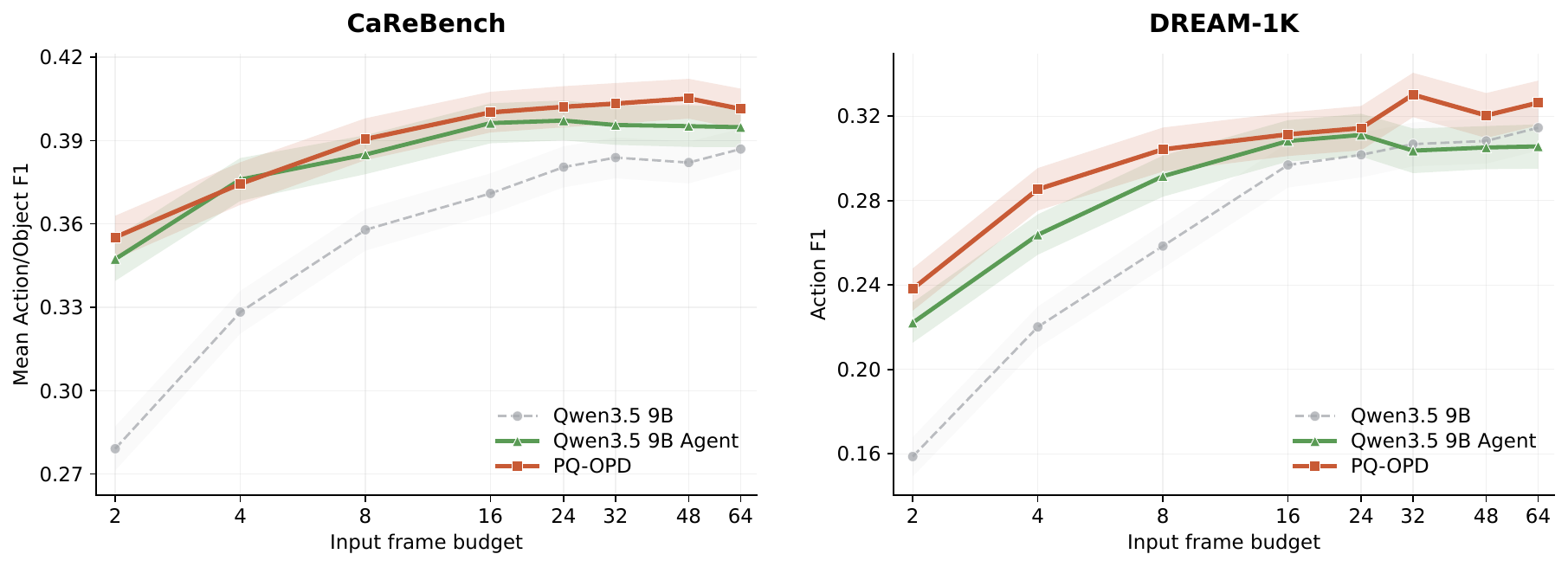}
  \caption{\textbf{Performance across input-frame budgets.} F1 under varying
  frame budgets, with a fixed evaluation cohort along each curve. CaReBench
  reports the arithmetic mean of Action F1 and Object F1; DREAM-1K reports
  event F1. Shaded regions show pointwise 95\% bootstrap intervals.}
  \label{fig:budget_curves}
\end{figure}

\method{} improves DREAM-1K F1 and CaReBench F1 across different
frame budgets (\cref{fig:budget_curves}). The advantage is also
clear under tight frame constraints, indicating that the learned agent uses
its available visual input more effectively.

The comparison with uniformly sampled direct input further highlights the
benefit of interactive observation. At small frame budgets, agentic models
achieve higher scores on both benchmarks despite operating
under the same input-frame limit. Across the full budget range, \method{}
consistently improves over the untrained agent, showing that PQ-OPD improves
captioning across different levels of visual input rather than relying on a
particular frame budget. Performance does not increase monotonically with the budget, indicating that
a larger visual budget alone does not guarantee better performance and that
the quality of the selected evidence also matters.
In particular, under the 32-frame cap matching fixed-input models, \method{}
reaches 40.3 / 33.0 on CaReBench / DREAM-1K, outperforming both uniform
sampling (38.4 / 30.7) and the untrained agent (39.6 / 30.4); detailed
inference statistics are provided in \cref{app:inference_cost}.
Qualitative examples of caption details and observation patterns are provided
in \cref{app:qualitative_examples}.

\begin{table}[t]
\centering
\caption{\textbf{Trajectory-swapping analysis (9B $\rightarrow$ 9B, 1{,}000 videos per benchmark).}
For cross-model swaps, the captioner writes the final description directly from
the frozen exploration trajectory collected by the other model, without issuing
further tool calls.}
\label{tab:trajectory_swap}
\small
\setlength{\tabcolsep}{8pt}
\renewcommand{\arraystretch}{1.10}
\begin{tabular}{@{}lcc@{}}
\toprule
Captioner $\leftarrow$ Trajectory & CaReBench F1 & DREAM-1K F1 \\
\midrule
Baseline agent $\leftarrow$ Baseline agent (\#3) & 39.8 & 31.3 \\
Baseline agent $\leftarrow$ \method{} & 40.3 & 32.6 \\
\method{} $\leftarrow$ Baseline agent & 40.3 & 32.2 \\
\rowcolor{gray!10}
\method{} $\leftarrow$ \method{} (\#11) & \textbf{40.6} & \textbf{33.3} \\
\bottomrule
\end{tabular}
\end{table}

\textbf{Disentangling exploration from caption generation.}
To separate gains in evidence acquisition from gains in caption writing, we
freeze the exploration trajectories of the untrained baseline agent
(\cref{tab:ablation}, \#3) and of \method{} (\#11) across all 1{,}000 videos of
each benchmark, and ask each model to write the caption directly from the
other's collected observations (\cref{tab:trajectory_swap}). Giving
\method{}'s trajectories to the baseline agent improves it from 39.8 / 31.3 to
40.3 / 32.6 ($+1.3$ of the $+2.0$ DREAM-1K gain), while under either trajectory
\method{} also outperforms the baseline agent ($+0.5$ / $+0.9$ and $+0.3$ /
$+0.7$).

\begin{figure}[t]
  \centering
  \includegraphics[width=\linewidth]{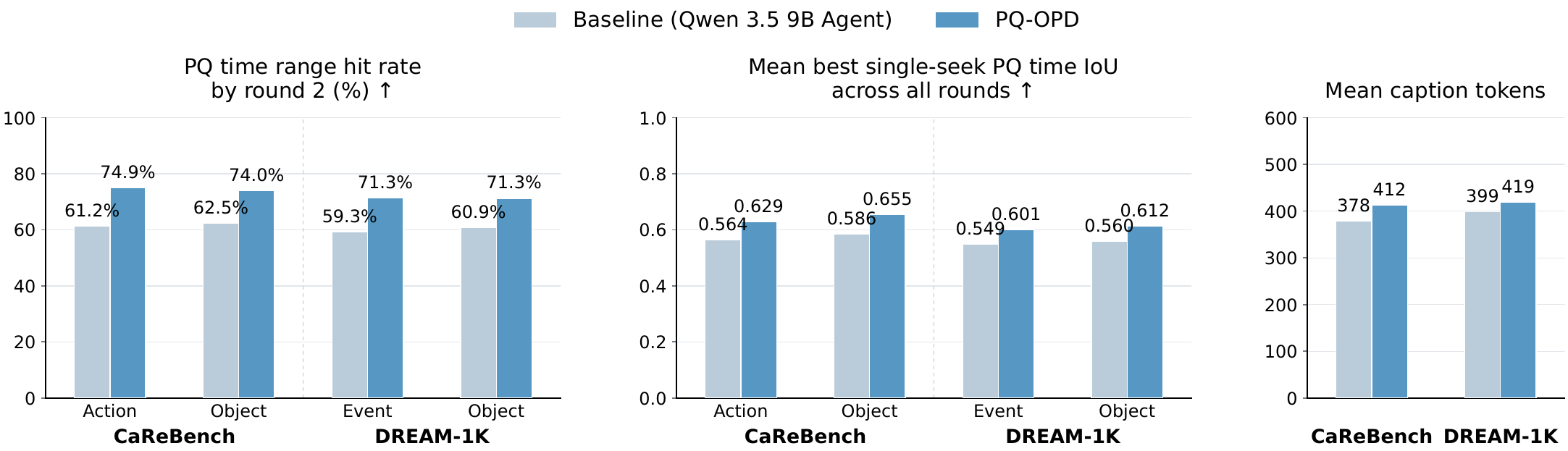}
  \caption{\textbf{Temporal exploration and caption length.}
  Left: PQ interval hit rate by round two (single-\texttt{seek} tIoU
  $\geq 0.5$). Middle: mean best single-\texttt{seek} tIoU across all rounds.
  Right: final-caption tokens. Temporal scores are averaged within each
  video and then across videos.}
  \label{fig:temporal_alignment}
\end{figure}

\textbf{Temporal exploration and caption length.}
We compare the base 9B agent and 9B $\rightarrow$ 9B PQ-OPD on
1{,}000 paired videos per benchmark. \Cref{fig:temporal_alignment} shows
that the trained agent reaches PQ-relevant temporal intervals earlier,
with Action/Object interval hit rates improving by 10.4--13.7 percentage
points across the two benchmarks. Its \texttt{seek} windows are also better
aligned with the PQ time intervals, increasing the mean best temporal intersection over union (tIoU) by
0.052--0.070. The final caption lengths remain similar,
increasing by only 9.0\% on CaReBench and 5.1\% on DREAM-1K. These results
suggest that the performance gains are empirically associated with a
change in exploration policy, rather than simply with
producing substantially different or longer final captions.

\begin{figure}[t]
  \centering
  \includegraphics[width=\linewidth]{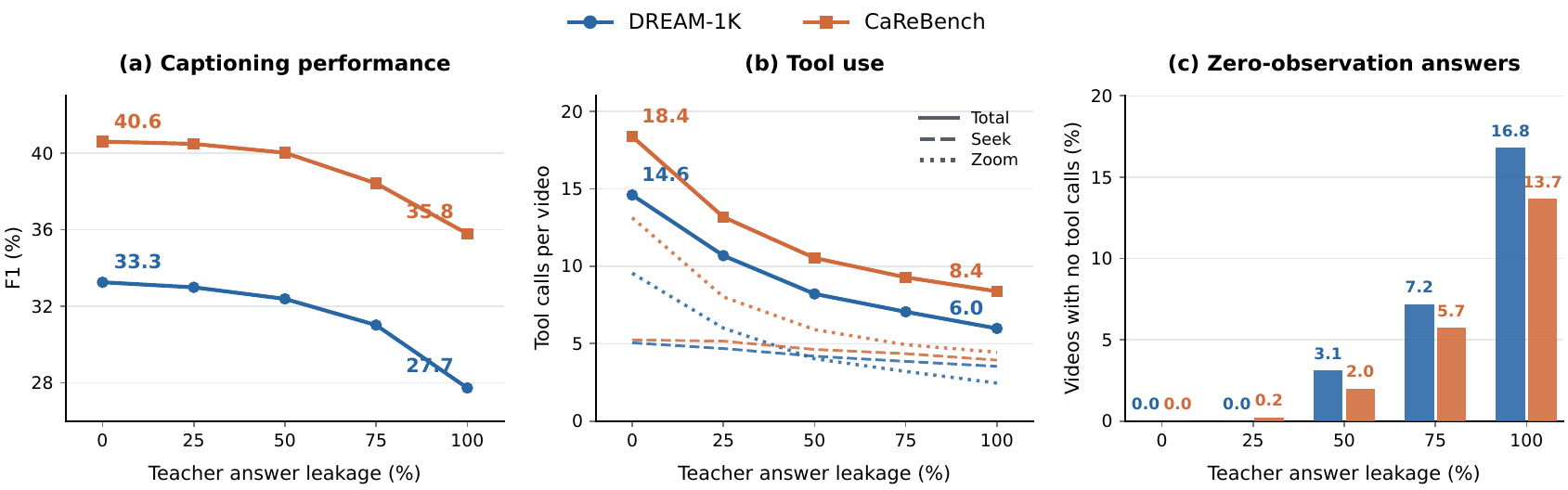}
  \caption{\textbf{Effect of answer leakage on PQ-OPD.} The rate of PQ
  answers leaked to the teacher during training varies from 0\% to 100\%;
  0\% is standard PQ-OPD. Both teacher and student are 9B models.
  (a) Captioning performance.
  (b) Mean tool calls per video.
  (c) Zero-observation answers: the student issues no tool calls and receives
  no visual input before producing a caption.}
  \label{fig:answer_disclosure}
\end{figure}

\textbf{Effect of answer leakage.}
We vary the fraction of PQ answers provided to the 9B teacher during
distillation and evaluate the resulting 9B student without questions or
answers at inference. \Cref{fig:answer_disclosure} shows that increasing
answer leakage is accompanied by lower captioning scores, fewer tool calls,
and more zero-observation answers. From 0\% to 100\% leakage, CaReBench F1
falls from 40.6 to 35.8 and DREAM-1K event F1 from 33.3 to 27.7. Average
tool calls per video decrease from 18.4 to 8.4 on CaReBench and from 14.6
to 6.0 on DREAM-1K, with a particularly pronounced reduction in zoom calls.
Meanwhile, the fraction of videos answered without any tool call rises from
zero to 13.7\% and 16.8\%, respectively. Since evaluation starts without
visual input, these answers are generated without observing the video.
The growing zero-observation fraction at higher
leakage suggests a failure to acquire visual evidence. Similar performance
degradation from answer-conditioned supervision is also observed in ViCuR~\citep{vicur2025}, which links answer-side privilege to shortcut learning.
These results support withholding PQ answers, which guides the
teacher toward relevant observations without supplying their outcomes.
Rewriting PQs at three increasing levels of wording leakage with fixed
timestamps (\cref{app:leakage_and_checklist}) further confirms that gains do
not come from leaked content: question-only captioning rises from 20.8 / 14.9
Action / Object F1 with our PQs to 52.7 / 52.9, whereas the trained students
drop to 37.3 / 29.4.

\section{Conclusion}
\label{sec:conclusion}

We introduced \method, an on-policy distillation framework that uses
training-only Perception Questions to guide a video captioning agent's
acquisition and description of visual evidence. A PQ-conditioned teacher
supervises student-generated interaction trajectories, while the student
receives no PQs at inference. Across five benchmarks, PQ-OPD improves caption
quality and downstream accuracy across model scales and frame budgets.
Ablations show benefits for both fixed-frame captioners and tool-enabled
agents, with on-policy training outperforming PQ trace off-policy distillation.
Temporal-coverage analysis further shows earlier, better-aligned exploration,
while answer leakage weakens visual evidence acquisition. These results
suggest that answer-free observation targets can be internalized as a
self-driven exploration policy.

\subsubsection*{Reproducibility Statement}
Training data construction, evaluation protocols, and evaluation metrics are
described in \cref{sec:experiments}. Complete prompt templates are
collected in \cref{app:prompts}, and all training hyperparameters are listed
in \cref{app:training_details}.
We promise to publicly release the PQ resource, generation code, and model weights.

\subsubsection*{AI Use Statement}
We used large language and vision-language models (Gemini) to generate
Perception Questions for training data construction (\cref{sec:setup}).
We also used language models as automated evaluators in the evaluation
process described in the paper, and to polish writing.
We take responsibility for all claims, numbers, and content in this paper.

\bibliography{iclr2027_conference}
\bibliographystyle{iclr2027_conference}

\appendix
\clearpage
\section{Training Details}
\label{app:training_details}

\Cref{tab:training_hyperparameters} lists the training configuration for the
Qwen3.5-9B student.

\begin{table}[!ht]
\centering
\caption{Training hyperparameters and configuration.}
\label{tab:training_hyperparameters}
\small
\renewcommand{\arraystretch}{1.12}
\begin{tabular}{@{}>{\raggedright\arraybackslash}p{0.40\linewidth}>{\raggedright\arraybackslash}p{0.56\linewidth}@{}}
\toprule
Parameter & Value \\
\midrule
\multicolumn{2}{@{}l}{\textit{Data and rollout}} \\
Training videos & 5{,}979 (\cref{sec:setup}) \\
Perception Questions & 43{,}477 \\
Maximum tool-image pixels & 460{,}800 \\
Unique prompts per refresh & 64 \\
Rollouts per prompt & 4 (256 trajectories per refresh) \\
\midrule
\multicolumn{2}{@{}l}{\textit{Optimization}} \\
Optimizer & AdamW \\
Adam coefficients & $\beta_1=0.9$, $\beta_2=0.95$ \\
Optimizer steps & 372 \\
Peak learning rate & $2\times10^{-6}$ \\
Learning-rate schedule & Cosine \\
Warmup ratio & 0.05 \\
Weight decay & 0.1 \\
Per-device batch size & 1 \\
Gradient accumulation steps & 8 \\
JSD $\beta$ & 0.5 \\
Temperature & 1.0 \\
\midrule
\multicolumn{2}{@{}l}{\textit{Memory and trainable components}} \\
Memory optimization & DeepSpeed ZeRO-3 \\
Vision encoder and aligner & Frozen \\
\bottomrule
\end{tabular}
\end{table}

\section{Additional Experimental Details and Analyses}
\label{app:additional_analyses}

\subsection{PQ Quality Control and Human Verification}
\label{app:pq_quality}

\paragraph{Automatic quality control.}
Each Perception Question is generated together with a concise answer and a
visual-evidence statement for its temporal interval (\cref{sec:setup},
\cref{app:pq_generation_prompt}). An automatic validator rejects and
regenerates outputs with invalid or mismatched time ranges, empty answers or
evidence descriptions, answers leaked into the question text, duplicate
questions, or event-premise clauses. All 43{,}477 PQs across the 5{,}979
training videos pass these checks, and the union of their temporal intervals
covers 97.6\% of each video's duration on average.

\paragraph{Human verification.}
Two human experts independently cross-verified 50 randomly sampled training
videos (344 PQs in total) against the source video frames. Among the questions
used in training, 99.7\% (343/344) ask about visually present content without
hallucinated premises, and 99.4\% (342/344) have temporal intervals covering
the target visual evidence. For the generated answers---which are excluded from
training---90.1\% (310/344) are also consistent with the video; because
\method{} never exposes answers during training, student learning is unaffected
by residual errors in the discarded answers.

\subsection{Generic Question OPD and Wording Leakage Controls}
\label{app:leakage_and_checklist}

\paragraph{Generic question specification.}
To match the question count ($\approx 7.3$ PQs per video) and type distribution
(33\% object, 36\% action, 31\% temporal) of our training set without any
video-specific object, action, or attribute terms, Generic Question OPD
(\cref{tab:ablation}, \#9) supplies every video of duration $T$ with the same
eight strictly neutral questions over uniformly partitioned quarter intervals:
\begin{enumerate}[leftmargin=*,nosep]
  \item What scene setting and visible elements appear between $0$ and $T/4$ seconds?
  \item What happens in the video between $0$ and $T/4$ seconds?
  \item What visual details and subject appearance are shown between $T/4$ and $T/2$ seconds?
  \item What happens in the video between $T/4$ and $T/2$ seconds?
  \item How are the scene layout, framing, or camera perspective shown between $T/2$ and $3T/4$ seconds?
  \item What happens in the video between $T/2$ and $3T/4$ seconds?
  \item What happens or changes in the video between $3T/4$ and $T$ seconds?
  \item How does the video progress overall from $0$ to $T$ seconds?
\end{enumerate}
Providing these generic questions to the frozen 9B teacher itself yields 40.0
CaReBench F1 and 32.9 DREAM-1K F1 (compared with 40.6 / 33.4 using
video-specific PQs), and distilling from them reaches 40.2 / 32.8.

\begin{table}[!ht]
\centering
\caption{\textbf{Wording content leakage controls.} All variants use the
9B $\rightarrow$ 9B recipe (372 steps) with identical timestamps.
``Question-only captioning'' reports Action / Object F1 of a
text-only model generating captions solely from the questions on a 1{,}000-video
training subset without visual input (for reference, an agent watching the
video reaches 48.7 / 55.8). ``Trained student'' reports standard PQ-free
evaluation on the full CaReBench and DREAM-1K benchmarks.}
\label{tab:wording_leakage}
\small
\setlength{\tabcolsep}{6pt}
\renewcommand{\arraystretch}{1.06}
\begin{tabular}{@{}lcccc@{}}
\toprule
\multirow{2}{*}{Teacher Questions}
& \multicolumn{2}{c}{Question-Only Captioning}
& \multicolumn{2}{c}{Trained Student} \\
\cmidrule(lr){2-3} \cmidrule(l){4-5}
& Action F1 & Object F1 & CaReBench F1 & DREAM-1K F1 \\
\midrule
\rowcolor{gray!10}
Standard PQs (\method{}) & 20.8 & 14.9 & \textbf{40.6} & \textbf{33.3} \\
L1 (leakier wording)     & 37.0 & 36.4 & 40.3 & 32.3 \\
L2 (more leaky wording)  & 43.6 & 46.5 & 39.9 & 32.5 \\
L3 (leakiest wording)    & 52.7 & 52.9 & 37.3 & 29.4 \\
\bottomrule
\end{tabular}
\end{table}

\paragraph{Wording content leakage gradient.}
To test whether \method{} benefits from video facts implicitly leaked through
question wording, we rewrite our training PQs at three increasing levels of
content leakage (L1--L3) while keeping identical timestamps, and train 9B
students with the exact \method{} recipe (372 steps). To quantify leaked
content, a text-only model writes a caption from the questions alone without
visual input on a 1{,}000-video training subset (\cref{tab:wording_leakage}).
Standard PQs yield only 20.8 / 14.9 Action / Object F1, whereas L1--L3 rise
from 37.0 / 36.4 to 52.7 / 52.9 (approaching an agent watching the video at
48.7 / 55.8). Rather than improving student performance, leakier questions
consistently degrade the distilled student down to 37.3 CaReBench F1 and 29.4
DREAM-1K F1, consistent with the explicit answer-leakage degradation in
\cref{fig:answer_disclosure}.

\subsection{Inference Cost Statistics}
\label{app:inference_cost}

\begin{table}[!ht]
\centering
\caption{\textbf{Per-video inference statistics on CaReBench and DREAM-1K.}
Evaluated across all 1{,}000 videos of each benchmark under the standard
16-round, 128-frame protocol (9B $\rightarrow$ 9B for \method{}). Visual tokens
count the image tokens of all returned observations (each capped at 460{,}800
pixels); latency is the median end-to-end wall-clock time per video under our
batched evaluation setting.}
\label{tab:inference_cost}
\small
\setlength{\tabcolsep}{5pt}
\renewcommand{\arraystretch}{1.06}
\resizebox{\linewidth}{!}{%
\begin{tabular}{@{}llccccc@{}}
\toprule
Benchmark & Model & Returned Frames & Tool Calls & Tool Rounds & Visual Tokens & Latency (Median) \\
\midrule
\multirow{2}{*}{CaReBench}
& Qwen3.5 9B Agent & 34.5 & 11.1 & 3.7 & 11.4K & 30\,s \\
& \method{}        & 54.6 & 18.4 & 4.1 & 17.6K & 41\,s \\
\midrule
\multirow{2}{*}{DREAM-1K}
& Qwen3.5 9B Agent & 31.3 &  9.6 & 3.3 & 12.7K & 31\,s \\
& \method{}        & 48.8 & 14.6 & 3.5 & 19.8K & 42\,s \\
\bottomrule
\end{tabular}%
}
\end{table}

\Cref{tab:inference_cost} reports per-video inference statistics under the
standard setting (\cref{sec:setup}), where \method{} uses roughly $1.5\times$
the returned frames and visual tokens of the untrained agent and $1.35\times$
its median latency. Crucially, \method{}'s gains do not come from viewing more
frames, but from selecting better ones: under identical frame caps
(\cref{fig:budget_curves}), \method{} reaches 39.1 / 30.4, 40.0 / 31.1, and
40.3 / 33.0 at 8, 16, and 32 frames on CaReBench / DREAM-1K, consistently
outperforming uniform sampling (35.8 / 25.8, 37.1 / 29.7, 38.4 / 30.7) and the
untrained agent (38.5 / 29.2, 39.6 / 30.8, 39.6 / 30.4).

\clearpage
\section{Prompt Templates}
\label{app:prompts}

This section collects the four prompt templates. Colored placeholders are
replaced by the video duration (in seconds, to three decimal places) or its
numbered Perception Questions. Typography and line wrapping are adapted for
readability; the prompt wording is preserved.

\subsection{Perception Question Generation}
\label{app:pq_generation_prompt}

The prompt template covers object, action, and temporal questions.
The generator receives the video and its duration; generated answers
and evidence are retained for quality control. Only questions reach the teacher.

\begin{promptbox}{PQ generation \hfill Instructions and coverage}
Watch the entire video and create a compact set of timestamped questions that will help another video model inspect the video and produce a complete, accurate description.

The video duration is \textcolor{promptaccent}{\texttt{\{duration:.3f\}}} seconds.

Generate approximately 6 to 10 questions in total. Choose the number according to the visual and temporal complexity of the video. Do not add redundant questions merely to reach a target count.

\textbf{Use three equally important question types:}

\textbf{1. {\small\ttfamily object}}\par Ask about concrete visible objects that are important for understanding the video, especially objects that are used, manipulated, exchanged, carried, or otherwise involved in the main activity.

\textbf{2. {\small\ttfamily action}}\par Ask about meaningful visible actions, movements, or interactions occurring within the stated interval.

\textbf{3. {\small\ttfamily temporal}}\par Ask about temporal order, action sequences, activity progression or completion, meaningful state changes, or repeated behavior across the stated interval.

Include all three question types, but do not follow a fixed numerical ratio. Allocate questions according to what is actually important in the video. Object, action, and temporal questions should be treated as equally valid ways to guide inspection rather than as separate generation stages.

\textbf{For temporal questions, prioritize:}

\begin{itemize}[leftmargin=1.25em,itemsep=1.5pt,topsep=2pt,parsep=0pt]
\item complete action chains;
\item chronological progression or completion of an activity;
\item meaningful changes in the state, position, or arrangement of a subject or object;
\item repeated or cyclic behavior.
\end{itemize}

At least one temporal question should cover the main progression of the video and span at least 60\% of its duration.

Avoid creating a static inventory of every visible object. Do not focus on clothing details, colors, background decoration, text, watermarks, lighting, camera movement, or other incidental content unless it is important to the main activity.

\textbf{Every question must:}

\begin{itemize}[leftmargin=1.25em,itemsep=1.5pt,topsep=2pt,parsep=0pt]
\item contain exactly one explicit time range written as {\small\ttfamily between X and Y seconds} or {\small\ttfamily from X to Y seconds};
\item be answerable from visible evidence within that interval;
\item leave the requested object identity, action, event sequence, or resulting state unknown;
\item use only the timestamp and, when necessary, a neutral visible anchor such as {\small\ttfamily the main subject}, {\small\ttfamily the seated person}, or {\small\ttfamily the main visible object};
\item contribute information useful for producing a detailed video description.
\end{itemize}

\textbf{Collectively, the questions should:}

\begin{itemize}[leftmargin=1.25em,itemsep=1.5pt,topsep=2pt,parsep=0pt]
\item cover the important content near the beginning, middle, and end of the video;
\item cover central subjects, objects, actions, interactions, and temporal progression;
\item avoid duplication and unnecessary detail.
\end{itemize}

\end{promptbox}

\clearpage
\begin{promptbox}{PQ generation \hfill Constraints and output format (continued)}
\textbf{Leakage restrictions:}

\begin{itemize}[leftmargin=1.25em,itemsep=1.5pt,topsep=2pt,parsep=0pt]
\item Do not reveal the answer in the question.
\item Do not state an observed event as the premise for another question.
\item Do not use premises such as {\small\ttfamily after the person picks up the cup}, {\small\ttfamily before the vehicle stops}, {\small\ttfamily following the action}, or {\small\ttfamily once the subject enters}.
\item Do not reveal concrete actions, object identities, or final states that the question is intended to ask about.
\item Do not use proper names or non-visual identities.
\item Do not ask why something happens or infer intent, causality, emotion, or hidden state.
\item Do not generate yes/no questions.
\item Do not duplicate or paraphrase another generated question.
\end{itemize}

\textbf{Preferred forms:}

\begin{itemize}[leftmargin=1.25em,itemsep=1.5pt,topsep=2pt,parsep=0pt]
\item {\small\ttfamily Which object is the person using between 2.0 and 4.0 seconds?}
\item {\small\ttfamily What important object is involved from 4.0 to 6.0 seconds?}
\item {\small\ttfamily What happens between 2.0 and 4.0 seconds?}
\item {\small\ttfamily What does the main subject do from 4.0 to 6.0 seconds?}
\item {\small\ttfamily What sequence of actions does the main subject perform from 0.0 to 12.0 seconds?}
\item {\small\ttfamily What happens first, next, and finally between 2.0 and 10.0 seconds?}
\item {\small\ttfamily How does the main subject's activity progress from 1.0 to 8.0 seconds?}
\item {\small\ttfamily How does the state of the main visible object change between 3.0 and 9.0 seconds?}
\item {\small\ttfamily What action pattern repeats, and how does it develop, from 0.0 to 15.0 seconds?}
\end{itemize}

For quality control, provide a concise answer and a short visual evidence statement for every question. Answers, evidence, type metadata, and temporal subtype metadata will not be shown to the evaluated model.

For a temporal question, assign one of these subtypes: {\small\ttfamily action\_chain}, {\small\ttfamily activity\_progression}, {\small\ttfamily state\_change}, or {\small\ttfamily repetition}. Object and action questions do not need a temporal subtype.

\textbf{Return JSON only:}

\begin{Verbatim}
{
  "questions": [
    {
      "question": "...?",
      "answer": "...",
      "evidence": "...",
      "start_time": 0.0,
      "end_time": 2.0,
      "type": "object|action|temporal",
      "temporal_subtype": null
    },
    {
      "question": "...?",
      "answer": "...",
      "evidence": "...",
      "start_time": 0.0,
      "end_time": 10.0,
      "type": "temporal",
      "temporal_subtype": "action_chain|activity_progression|state_change|repetition"
    }
  ]
}
\end{Verbatim}
\end{promptbox}

\clearpage
\subsection{Agent System Prompt}
\label{app:agent_system_prompt}

The student uses this system prompt for tool-interaction rollouts. The teacher
scores the same history under this setting with a PQ-conditioned user query.

\begin{promptbox}{Agent \hfill System message}
You are a video-analysis agent. The source video is \textcolor{promptaccent}{\texttt{\{duration:.3f\}}} seconds long and is available through the provided tools.

\textbf{You must use the tools to explore the video before answering:}

seek(t\_start, t\_end) takes times in seconds and returns up to 8 representative frames from that interval.\\zoom(timestamp, bbox) returns a crop from one frame; bbox is [x1, y1, x2, y2] in 0-1000 coordinates.

You may call multiple tools in the same turn. In the final answer, do not mention the tools, hidden-input setup, or these instructions.

\end{promptbox}

\subsection{Student User Prompt}
\label{app:student_prompt}

The student receives the generic instruction during training and inference.

\begin{promptbox}{Student \hfill User message}
Describe the video in detail.

\end{promptbox}

\subsection{PQ-Conditioned Teacher User Prompt}
\label{app:teacher_prompt}

The frozen teacher receives numbered PQs while sharing the student's actions
and observations. Answers and reference captions are excluded.

\begin{promptbox}{Teacher \hfill User message · Training only}
Describe the video in detail. Include all visually supported information needed to answer the questions below. Do not mention or repeat the questions, and do not answer them one by one.

Questions:\\1. \textcolor{promptaccent}{\texttt{\{q\_1\}}}\\2. \textcolor{promptaccent}{\texttt{\{q\_2\}}}\\...\\K. \textcolor{promptaccent}{\texttt{\{q\_K\}}}

Output only the video description.

\end{promptbox}

\clearpage
\section{Qualitative Examples}
\label{app:qualitative_examples}

We compare selected caption excerpts and observation patterns from the
non-agentic Qwen3.5~9B model, the untrained agent, and \method{} on two
DREAM-1K videos. Both agents operate without PQs at inference.

\begin{figure}[!ht]
  \centering
  \includegraphics[width=\linewidth]{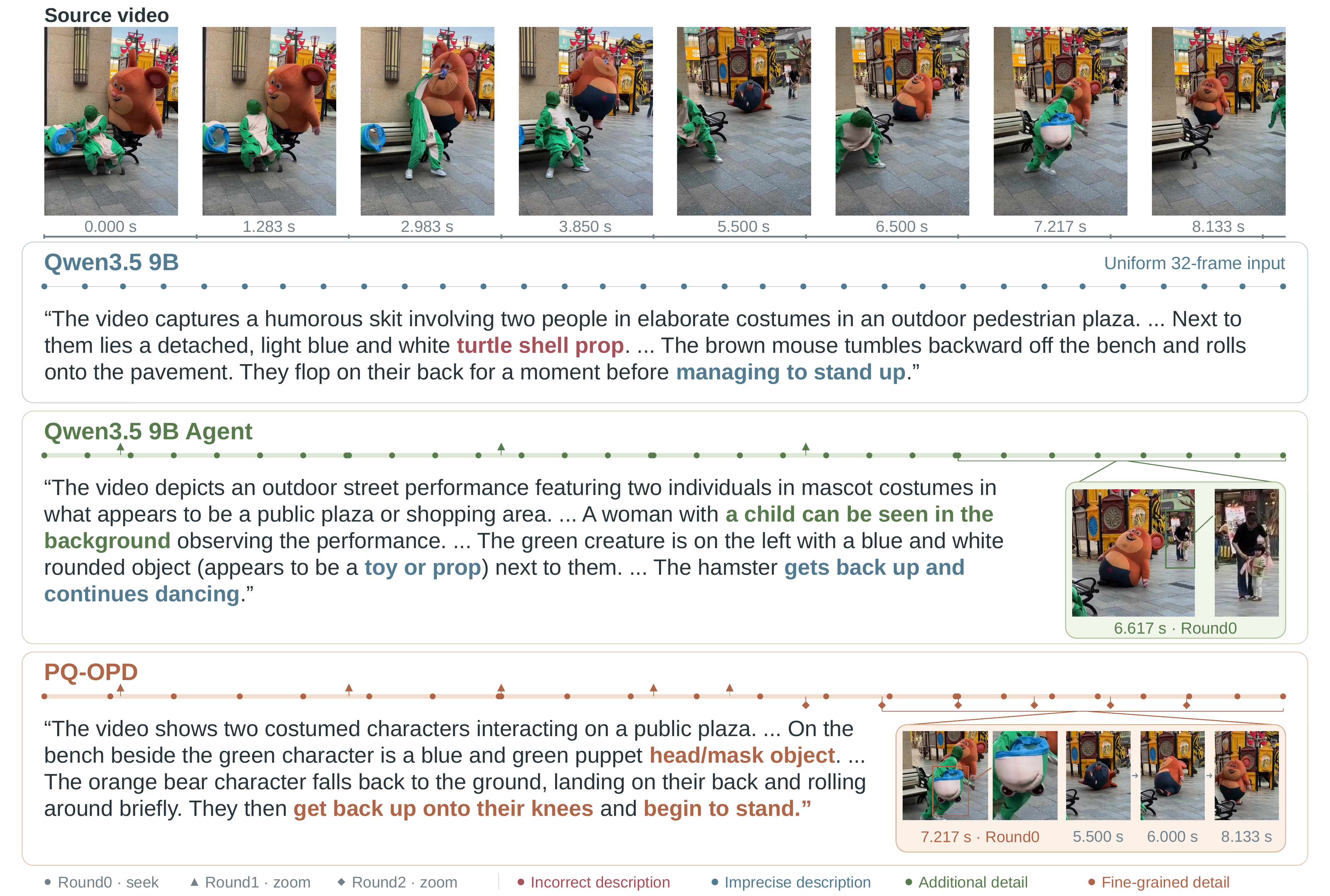}
  \caption{\textbf{Qualitative comparison on DREAM-1K.}
  Qwen3.5-9B receives 32 uniformly sampled frames, while Qwen3.5-9B Agent
  and \method{} each obtain 35 visual observations through tool use.
  Circles denote the initial \texttt{seek} observations, triangles denote
  first-round single-frame inspections, and diamonds denote second-round
  inspections. The enlarged panels show cropped regions for easier visual
  comparison. The examples highlight differences in object identification,
  background details, and the fallen character's recovery motion.}
  \label{fig:qualitative_sample}
\end{figure}

\paragraph{Costume interaction.} \Cref{fig:qualitative_sample} compares
captions and observation patterns on a DREAM-1K video. Both agentic models
receive 35 visual observations: the base agent uses 32 initial seek frames
and 3 subsequent single-frame zoom crops, whereas \method{} uses 24 seek
frames and 11 zoom crops. The base agent captures a background child omitted by the
fixed-frame caption. \method{} describes the detached costume head more
specifically than the base agent's ``toy or prop'' and further captures the
fallen character getting onto its knees and beginning to stand. Despite
identical returned-image counts, the two agents capture different details.

\begin{figure}[!ht]
  \centering
  \includegraphics[width=\linewidth]{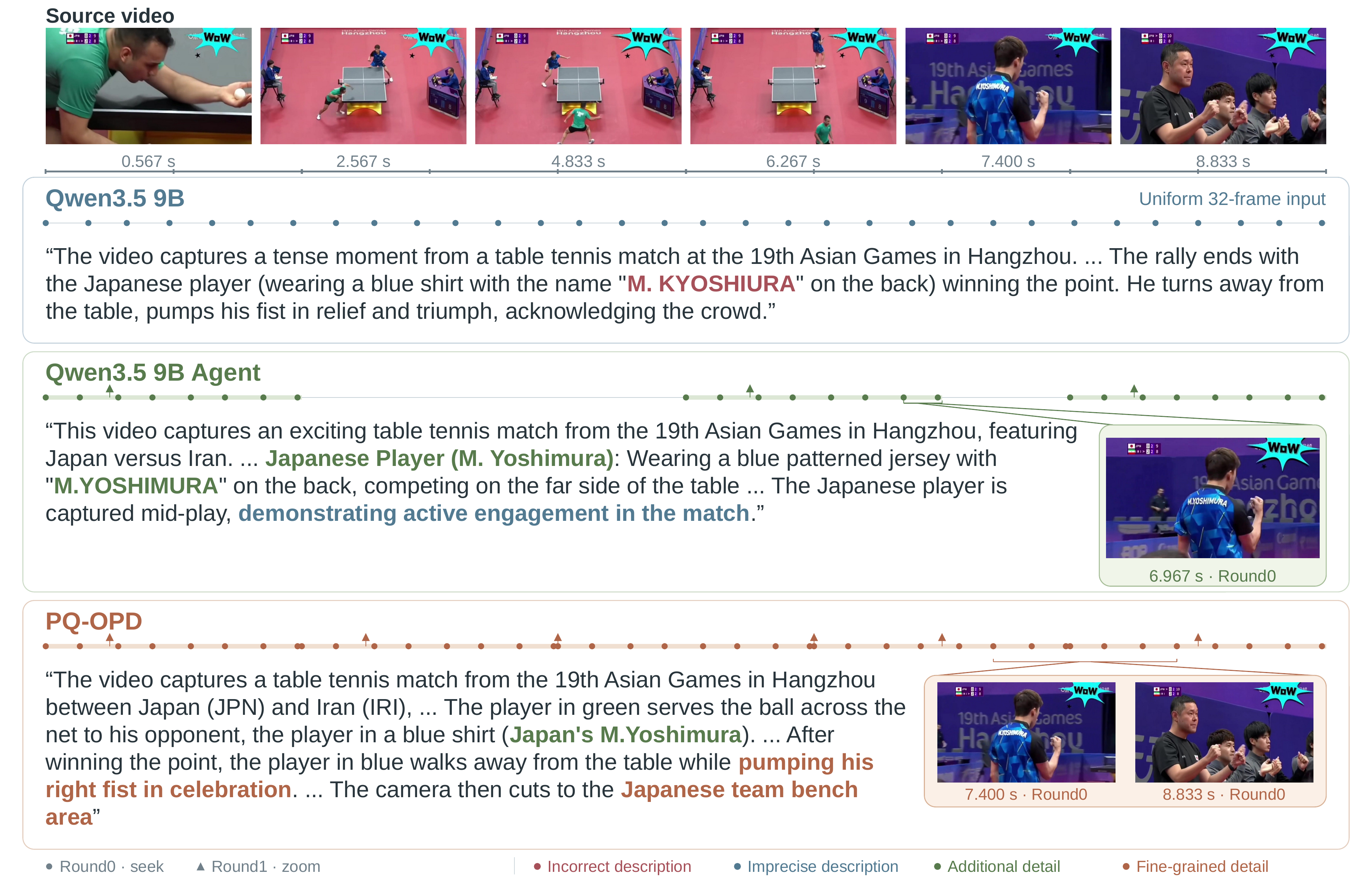}
  \caption{\textbf{Player identity and action details on DREAM-1K.}
  Qwen3.5-9B receives 32 uniformly sampled frames, while the two agentic
  models acquire observations through \texttt{seek} and \texttt{zoom}.
  Circles denote the initial \texttt{seek} observations, and triangles denote
  first-round \texttt{zoom} inspections. The comparison highlights jersey-name
  recognition and the description of the player's fist-pump celebration and
  the subsequent cut to the team bench. Enlarged panels show the relevant
  visual evidence.}
  \label{fig:qualitative_sample2}
\end{figure}

\clearpage
\paragraph{Table-tennis match.} \Cref{fig:qualitative_sample2} compares
identity recognition and action details in a table-tennis match. The base
agent correctly reads ``M.YOSHIMURA'' on the player's jersey, whereas the
non-agentic caption misspells the name. \method{} preserves the correct
identity and further describes the player pumping his right fist before the
video cuts to the team bench.

\section{Limitations}

\method relies on an external video model to generate Perception
Questions offline, introducing additional data-construction cost
and potential generator bias. Our evaluation also relies substantially
on model-based judges, including replacement judges for benchmarks
whose original evaluators are no longer publicly accessible;
absolute scores may therefore vary with the evaluator and prompt.
At inference, the trained agent requires no PQs, but iterative \texttt{seek}/\texttt{zoom}
calls are more expensive than a single fixed-frame forward pass.
Finally, our experiments focus on a specific 
\texttt{seek}/\texttt{zoom} interface and Qwen3.5 backbones.
Whether the learned observation policy transfers to other tool interfaces,
and other model families remains an open question.

\end{document}